\documentclass[letterpaper,10pt,conference]{ieeeconf}

\IEEEoverridecommandlockouts
\usepackage{amsmath,amssymb}
\usepackage{graphicx}
\usepackage{booktabs}
\usepackage{makecell}
\usepackage{cite}
\usepackage{hyperref}
\usepackage{placeins}
\usepackage{iftex}

\ifPDFTeX
\fi

\title{\LARGE\bf
Learning Language-Conditioned Traversability Representations for Adaptive Visual Navigation
}

\author{Senda Chen, Changxu Cheng$^{*}$, Fangdi Li, Tao Wang, Wuyue Zhao%
\thanks{$^{*}$Corresponding author.}\\
Uni-Ubi\\
{\tt\small \{sendachen586,ccx0127\}@gmail.com}%
}

\begin{document}

\maketitle
\thispagestyle{empty}
\pagestyle{empty}

\begin{abstract}
    Traversability is essential for visual navigation but varies with robot capabilities and user preferences. Conventional pipelines often rely on explicit costmaps or segmentation masks with predefined criteria, requiring hand-crafted rules and careful tuning. Moreover, viewpoint-dependent segmentation masks complicate asynchronous planning under perception latency.
    We present LaTraNav, a framework that learns language-conditioned traversability representations for adaptive visual navigation. Its asynchronous architecture combines a slow vision-language model that produces latent representations of traversability and navigation goals, with a fast flow-matching planner conditioned on these representations.
    To train the system, we develop a simulation-based data generation pipeline with controllable trajectories, producing observations paired with language instructions, traversability maps, goal locations, and diverse trajectories. Photorealistic image translation further enhances visual realism.
    Evaluations on datasets from multiple sources demonstrate effective language-guided traversability segmentation and goal localization by the slow VLM, alongside adaptive pixel-space path planning by the fast planner. Latent conditioning improves planning performance over explicit segmentation masks, while asynchronous scheduling increases the path-update rate by $6.05\times$ at the same semantic-update rate.
\end{abstract}

\section{Introduction}

Traversability estimation is fundamental to visual navigation: it identifies the regions a robot can navigate through to reach a goal.
Costmaps and semantic segmentation have enabled planners to account for terrain properties beyond geometric obstacle avoidance. ViPlanner~\cite{ref37} combines geometric and semantic observations in a learned local planner, using differentiable semantic costmaps for training. Wild Visual Navigation~\cite{ref38} learns visual traversability through online self-supervision. These advances demonstrate the value of scene semantics for route selection in environments where geometric free space alone is insufficient.

\begin{figure}[!t]
\centering
\includegraphics[width=\columnwidth]{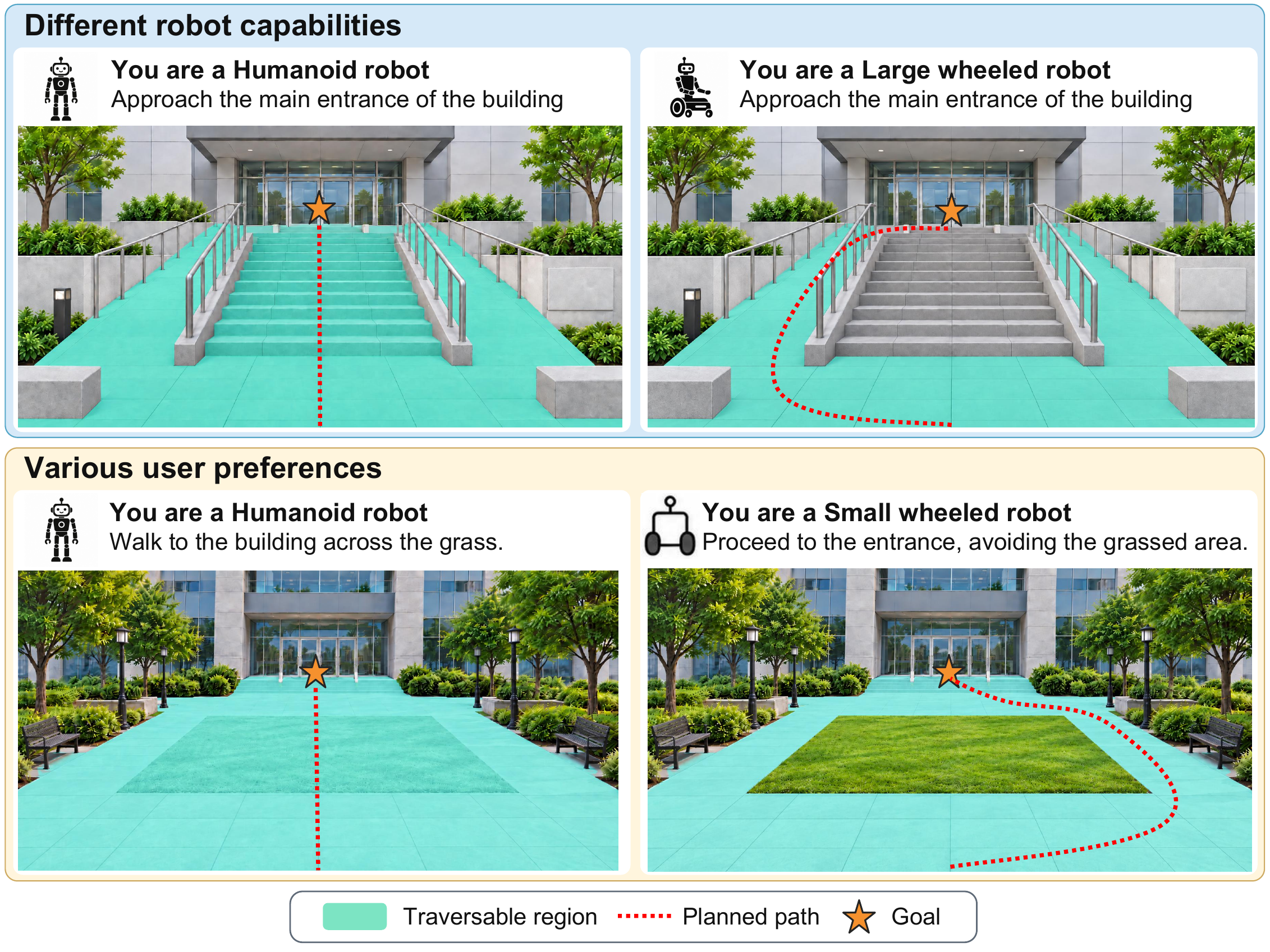}
\caption{Traversability depends on robot capabilities and language-specified requirements. With the scene and goal fixed, these conditions determine traversable regions and feasible paths, as illustrated by stairs versus ramps and grass crossings versus paved passages.}
\label{fig:task_overview}
\end{figure}

However, traversability depends on robot capabilities and user requirements. Stairs may be accessible to a humanoid but unsuitable for a wheeled robot, while an instruction to stay on paved surfaces can exclude otherwise traversable grass (Fig.~\ref{fig:task_overview}).
Existing planning pipelines face two practical challenges.
First, pipelines based on predefined terrain classes and hand-assigned costs rely on heuristic rules and manual cost tuning to encode these conditions~\cite{ref37,ref6}. PACER~\cite{ref6} improves preference adaptation through learned costmap generation, but retains an explicit costmap as the planning interface.
Second, VLM inference latency complicates asynchronous planning when traversability is represented by traversability masks. A mask refers to the viewpoint at which it was predicted, so reusing it after camera motion requires spatial alignment or regeneration. This makes it difficult to update paths more frequently than semantic updates.

Referring expression segmentation offers a starting point for more flexible traversability understanding. LAVT~\cite{ref39} grounded natural-language descriptions in pixel-level masks. Several methods~\cite{ref12,ref13,ref40,wang2026alto} further connect the reasoning capabilities of large vision-language models (VLMs) to segmentation through language-conditioned hidden states. These developments suggest that robot capabilities and traversal requirements can be expressed as text to identify task-specific regions. The hidden states for segmentation also provide a potential planning bridge, carrying region semantics beyond the decoded mask. This motivates learning a traversability representation that supports both language-guided region understanding and frequent path generation.

We present LaTraNav (Language-Conditioned Traversability Navigation), a dual-system framework that couples a slow VLM with a fast flow-matching planner through a latent traversability representation (Trav. Rep.).
Built on UFO~\cite{ref13}, the slow VLM interprets an RGB observation and an instruction describing the agent, destination, and traversal requirements. It predicts a goal point in normalized image coordinates and a traversability mask, retaining the mask-token hidden states as Trav. Rep. The flow-matching planner combines this representation with current visual features and the goal to generate an image-space path through conditional flow matching~\cite{ref11}.
Pixel-level segmentation supervision grounds the representation in agent- and language-conditioned traversable regions, while path supervision jointly adapts it to planning. This Trav. Rep. interface supports asynchronous execution: the flow-matching planner reuses the goal and Trav. Rep. between semantic updates and incorporates current visual observations, allowing path updates without repeating VLM inference at every planning step.

To train both systems, we develop a controllable data-construction pipeline with controllable scene geometry, agent conditions, and reference paths. The pipeline pairs visual observations with language instructions, traversability masks, goals, and paths, including different agent conditions within shared scenes. Photorealistic image generation expands appearance coverage, while existing visual datasets are incorporated through a unified annotation process. Together, these sources provide aligned supervision for agent- and language-conditioned region understanding and path generation across robot platforms.

Experiments across multiple data sources evaluate region, goal, and route feasibility predictions by the slow VLM, together with adaptive path generation by the flow-matching planner. On 13,580 evaluation samples, LaTraNav achieves a traversability IoU of 0.8817 and a joint goal-arrival and region-adherence success rate of 63.95\% across all ground-truth feasible tasks. Conditioning on Trav. Rep. improves planning over explicit masks. Asynchronous execution increases the path-update rate from 1.21 to 7.32 Hz ($6.05\times$) while the semantic-update rate remains at 1.21 Hz. Ablations further demonstrate the benefits of pixel-level segmentation supervision and joint optimization for the learned representation.

The main contributions are summarized as follows.

\begin{itemize}
\item A dual-system architecture that connects a slow VLM and a flow-matching planner through goals and Trav. Rep., supporting asynchronous language-guided path generation.
\item An agent- and language-conditioned traversability representation jointly optimized by pixel-level segmentation supervision and path learning, linking robot capabilities and task requirements to route selection.
\item A controllable data-construction pipeline combining simulation, photorealistic image generation, and adapted visual sources to provide aligned language, region, goal, and path supervision across robot platforms.
\end{itemize}

\section{Related Work}
\subsection{Visual Navigation}

Visual navigation connects observations to goal-directed motion. ViNT~\cite{ref19} and NoMaD~\cite{ref20} learn transferable policies, while NavDP~\cite{cai2025navdp,yang2026x} and SIDP~\cite{zhang2026self} develop diffusion policies for robust trajectory generation. Language-guided systems ground instructions through pretrained models~\cite{ref2}, vision-language maps~\cite{ref5}, or semantic frontier selection~\cite{ref16}. PixNav~\cite{ref4} further connects goal localization in normalized image coordinates to navigation. These methods identify and approach destinations; our problem additionally asks which image regions are traversable under the specified agent and route constraint.

Traversability-aware navigation incorporates terrain semantics, robot capabilities, and task requirements. ViPlanner~\cite{ref37} learns from geometric and semantic observations; Wild Visual Navigation~\cite{ref38} and VLM-GroNav~\cite{ref15} adapt traversability through interaction. ExAug~\cite{ref14} conditions policies on robot geometry, while CAT~\cite{ref3} injects platform limits into dense spatial features. SwarmDiffusion~\cite{swarmdiffusion} jointly predicts traversability and trajectories from RGB and embodiment state. PACER~\cite{ref6} and NORM-Nav~\cite{ref7} encode preferences or language requirements as planning costs, and CapNav~\cite{capnav} evaluates capability-conditioned feasibility and route validity from tour videos and navigation graphs. LaTraNav instead couples free-form route constraints with agent- and language-conditioned region prediction and image-space planning from a single view.

Hierarchical navigation also depends on the interface between semantics and motion. NaVILA~\cite{ref8} passes spatial instructions to locomotion, and VAMOS~\cite{ref9} scores image-space paths with an embodiment-specific affordance model. VENTURA~\cite{ref33} converts diffusion-generated path masks into trajectories, whereas PathPainter~\cite{ref10} predicts goals and traversability in bird's-eye view. CATNAV~\cite{catnav} caches VLM-derived, embodiment-aware costmaps for repeated trajectory selection. LaTraNav uses a goal and learned Trav. Rep. as its interface: pixel-level segmentation supervision and path supervision ground the VLM state for a flow-matching planner, enabling agent- and language-conditioned image-space planning and asynchronous path updates.

\subsection{Referring Expression Segmentation}

Referring expression segmentation (RES) identifies image regions described by language. LAVT~\cite{ref39} integrates language into visual encoding, and LSeg~\cite{ref18} aligns visual features with text embeddings for open-vocabulary labels. LISA~\cite{ref12} handles queries involving implicit intent and reasoning through segmentation-token hidden states. These advances allow traversability targets to vary with descriptions of agent capabilities and permitted regions.

HiMTok~\cite{ref40} represents masks as hierarchical token sequences, while UFO~\cite{ref13} retrieves regions by matching mask-token embeddings with image features. LaTraNav builds on this interface and exposes mask-token hidden states as Trav. Rep. to its flow-matching planner. Pixel-level segmentation supervision grounds the representation in agent- and language-conditioned traversable regions, and path supervision adapts it to path generation. The resulting Trav. Rep. interface connects RES to reusable conditioning for asynchronous planning.

\section{Method}

Given an egocentric RGB observation $I$ and instruction $\ell$, LaTraNav predicts a local image-space path $P=(p_1,\ldots,p_K)$ within the visible scene. The instruction specifies the agent, i.e., the robot embodiment, destination, and traversal constraints. The destination is the semantic target, its goal point $g$ is expressed in normalized image coordinates, and the path endpoint $p_K$ is the final generated point. Traversability denotes region semantics conditioned on the image and instruction.

The task predicts agent-conditioned route feasibility $y$, goal point $g$, and an agent- and language-conditioned traversable region. Trav. Rep. $Z$ encodes this traversability and connects the VLM $\mathcal F_\theta$ to the flow-matching planner $\mathcal G_\phi$:

\begin{equation}
\begin{aligned}
(\hat y,\hat g,Z)&=\mathcal F_\theta(I,\ell),\\
\hat M&=\mathcal D_\theta(I,Z),\\
H&=A_\psi(Z),\\
\hat P&=\mathcal G_\phi(I,s,\hat g,H,\epsilon),
\quad \epsilon\sim\mathcal N(0,\mathbf I),\quad \hat y=1.
\end{aligned}
\end{equation}

\begin{figure*}[t]
\centering
\includegraphics[width=\textwidth]{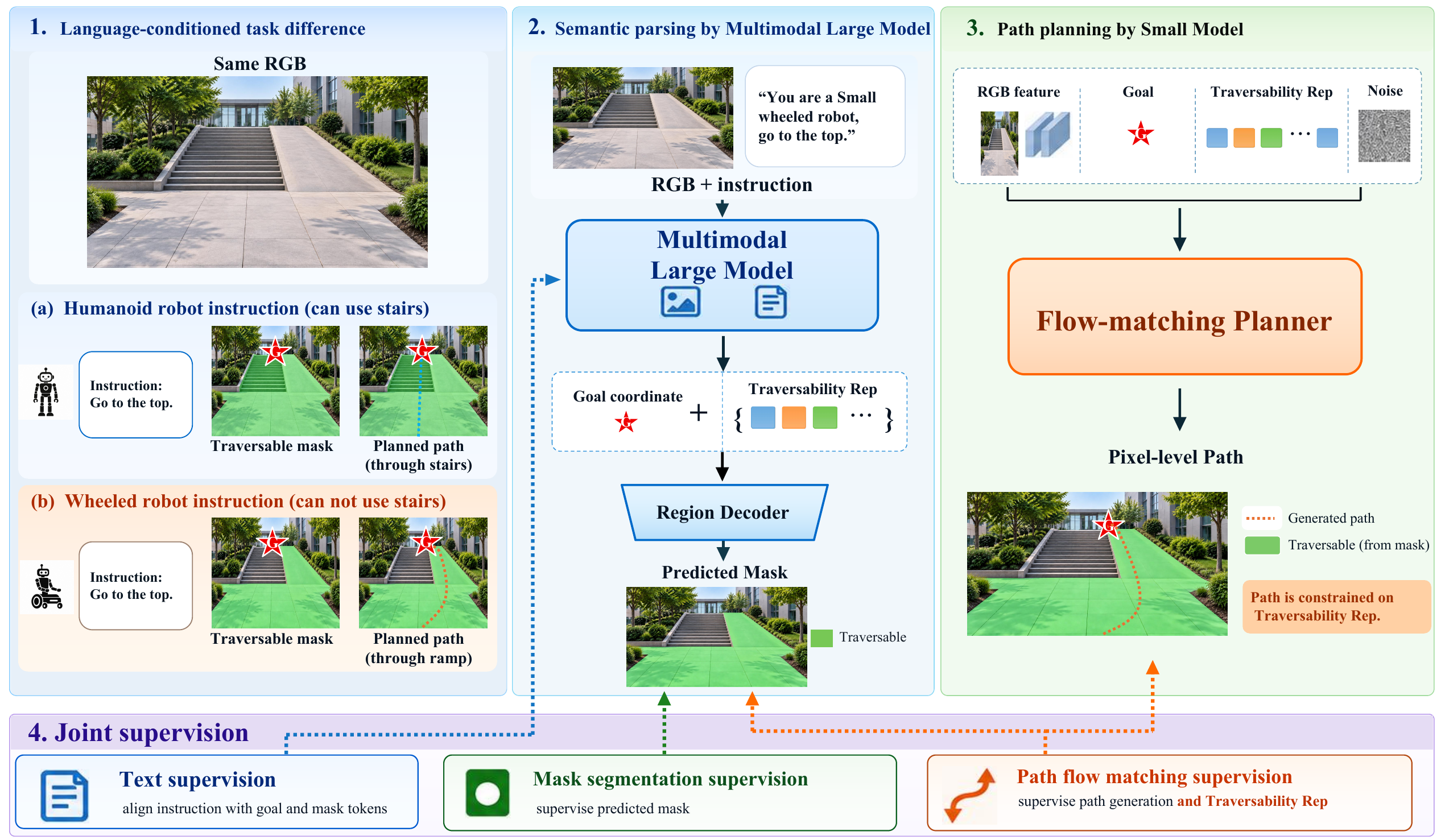}
\caption{LaTraNav couples a VLM with a flow-matching planner. Agent conditions or language constraints yield different traversable regions and paths in the same scene. The VLM predicts the goal, traversability mask, and Trav. Rep. $Z$. The planner combines visual features, the goal, and adapted planning tokens $H=A_\psi(Z)$ to generate an image-space path. Text supervision trains goal and agent-feasibility outputs, while pixel-level segmentation supervision and path supervision jointly shape $Z$.}
\label{fig:framework}
\end{figure*}

Here, $s$ is a path anchor in normalized image coordinates, given by the first reference-path point during training and $(0.5,1)$ at inference. $\mathcal D_\theta$ is the mask decoder, $A_\psi$ is a feature adapter, and $\epsilon$ is initial noise. The VLM generates $y$, $g$, and mask tokens whose hidden states form $Z$. Decoding $Z$ yields the traversability mask $\hat M$, while $A_\psi$ maps $Z$ to planner tokens $H$. The flow-matching planner generates a path when $\hat y=1$; otherwise, the VLM returns a refusal. Fig.~\ref{fig:framework} illustrates the architecture.

This interface separates VLM inference from path generation. Until the next semantic update, the flow-matching planner combines the cached goal and Trav. Rep. with current visual features. Each semantic update replaces both cached conditions.

\subsection{Learning Traversability Representations for Planning}

Traversability-mask prediction identifies traversable image locations, whereas path generation organizes them toward the goal. The tasks share spatial semantics but have different targets. Trav. Rep. is therefore shared by mask decoding and path generation, allowing traversability-mask annotations to ground its semantics and path supervision to adapt it to route selection.

Using UFO's segmentation interface\cite{ref13}, the VLM generates mask tokens conditioned on $I$ and $\ell$, whose hidden states form $Z\in\mathbb R^{M\times d}$. Goal coordinates are generated in the same response. UFO's decoder associates $Z$ with image features to produce the traversability mask, providing pixel-level segmentation supervision. A normalization and multilayer-perceptron adapter maps $Z$ to adapted planning tokens $H$.

The flow-matching planner consumes $H$, and its path loss backpropagates through the adapter to parameters producing $Z$. Pixel-level segmentation supervision grounds the Trav. Rep. in traversable regions, while path supervision adapts it to route selection, without passing a thresholded mask to the planner.

\subsection{Learning Region-Conditioned Path Distributions}

The flow-matching planner combines current visual features with $H$, the goal, and path anchor $s$. A frozen DINOv2 encoder\cite{ref24} supplies visual features, while $H$ conveys the traversability encoded by $Z$. Following the grouped attention design of $\pi_0$\cite{ref25}, noisy path tokens attend to one another and these conditions; the conditioning tokens do not attend to the noisy path.

Path generation follows conditional flow matching\cite{ref11}. A conditional vector field transforms Gaussian noise into consecutive-point increments. The path objective $\mathcal L_{\mathrm{path}}$ is the mean-squared error between the predicted field and target velocity along a linear noise-to-data interpolation. At inference, Euler integration generates increments accumulated relative to $s$. The goal and $H$ condition every update on the destination and permitted regions.

\subsection{Joint Optimization}

LaTraNav jointly optimizes the VLM and flow-matching planner through the same mask-token hidden states. Samples contain images, instructions, goals, traversability-mask annotations, and reference paths. Paths are resampled by arc length in normalized image coordinates, with the first point providing $s$. An incompatible route produces a refusal and its reason; the agent-conditioned route feasibility label follows the instruction.

The VLM response sequence is supervised by autoregressive cross-entropy $\mathcal L_{\mathrm{text}}$, and traversability-mask predictions are supervised by the segmentation loss $\mathcal L_{\mathrm{seg}}$, comprising focal and Dice losses. The joint objective is

\begin{equation}
\mathcal L
=\lambda_{\mathrm{text}}\mathcal L_{\mathrm{text}}
+\lambda_{\mathrm{seg}}\mathcal L_{\mathrm{seg}}
+\lambda_{\mathrm{path}}\mathcal L_{\mathrm{path}}.
\end{equation}

Here, $\lambda$ denotes the loss weights. Segmentation and path losses are computed only for compatible samples. During training, mask-token hidden states are extracted with teacher forcing, and the reference goal $g^*$ conditions the flow-matching planner. Goal localization is therefore learned through text supervision, while the Trav. Rep. receives both pixel-level segmentation supervision and path supervision. Path gradients reach the VLM through the Trav. Rep. interface, rather than through the discrete generation of goal coordinates.

At inference, both goal coordinates and mask tokens are generated autoregressively by the VLM. Paths are then sampled by the flow-matching planner using the predicted goal and the corresponding Trav. Rep.

\section{Controllable Data Construction}
\label{sec:data}

Training LaTraNav requires aligned instruction, traversability-mask, goal, reference-path, and agent-conditioned route feasibility annotations. The controllable data-construction pipeline combines a simulation branch with adaptation of existing visual sources. The former allows route geometry and robot capabilities to be varied within shared scenes, while the latter broadens scene and appearance coverage. Both are converted to the same planning-task format for training and testing.

\subsection{Simulation-Based Controllable Generation}

\begin{figure}[t]
    \centering
    \includegraphics[width=\columnwidth]{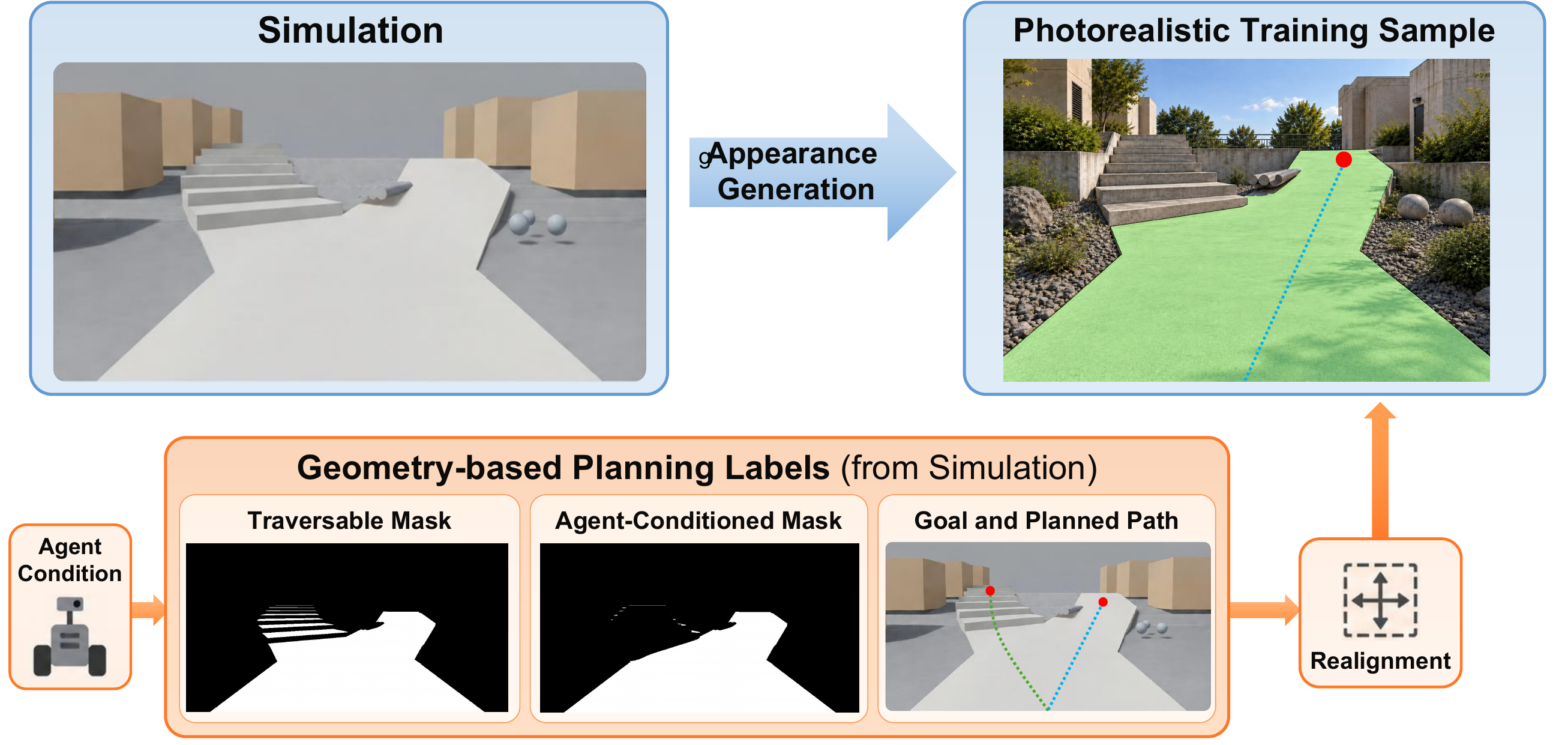}
\caption{Controllable construction of photorealistic planning samples. Simulation geometry provides agent- and language-conditioned traversable regions, goals, and reference paths. Photorealistic image generation changes visual style, after which the traversability-mask annotations are realigned to the generated image.}
    \label{fig:simulation_pipeline}
\end{figure}

Figure~\ref{fig:simulation_pipeline} summarizes the simulation branch. Simulated scenes vary road topology and obstacle placement to create direct traversal, obstacle-avoidance, and branch-selection tasks. For each scene, agent dimensions and safety margins determine traversable regions and A* reference paths in a metric bird's-eye-view grid. Isaac Sim camera geometry projects this supervision into the image. Holding scene geometry and appearance fixed while changing the agent produces corresponding traversability-mask and agent-conditioned route feasibility annotations for humanoid robots, large wheeled robots, and small wheeled robots. Instructions are generated from the corresponding scene and agent labels. This construction allows route requirements to be controlled through scene geometry and robot capabilities.

To expand appearance coverage, Qwen-Image-Edit\cite{ref41} performs photorealistic image generation conditioned on simulated RGB, the union of agent-specific traversability masks, and scene descriptions. The union retains candidate branches available to different agents, while each agent retains its own traversability-mask annotation. Mask2Former\cite{ref27} and SAM 3\cite{ref26} realign these annotations with the generated images. The Small Alley subset additionally provides narrow-passage imagery generated with GPT-Image-2\cite{ref35}.

\subsection{Adaptation of Existing Visual Sources}

Existing visual sources provide complementary scene coverage. ADE20K\cite{ref29}, FPV Crosswalk\cite{ref30}, CARLA\cite{ref31}, RoomTour3D\cite{ref32} and EgoWalk\cite{ref22} are processed through a common annotation pipeline. Usable traversability masks are retained, while missing or unsuitable masks are annotated or corrected with Mask2Former\cite{ref27} and SAM 3\cite{ref26}. Qwen3.8-27B\cite{ref28} selects goals from candidate locations, and a region-constrained planner generates reference paths. The image, goal, reference path, and agent description then condition instruction generation and agent-conditioned route feasibility annotation.

\subsection{Unified Format and Dataset Composition}

Samples from both branches share RGB, agent description, instruction, traversability mask, goal, reference path, and agent-conditioned route feasibility labels. The first point of the reference path provides its path anchor in normalized image coordinates. Existing scene annotations are thereby extended to agent- and language-conditioned planning supervision, while the simulation branch provides shared-scene agent variations. The resulting dataset contains 639,343 training samples and 13,580 test samples across the eight source groups in Table~\ref{tab:eval_data}. The training split is shared by LaTraNav and the independent cascade. Of the test samples, 11,768 are feasible and 1,812 are infeasible for the specified agent and route. The source composition covers indoor layouts, outdoor walking, crossings, simulated routes, and generated passages.

\begin{table}[t]
\caption{Training and test set composition}
\label{tab:eval_data}
\centering
\scriptsize
\setlength{\tabcolsep}{3pt}
\begin{tabular*}{\columnwidth}{@{\extracolsep{\fill}}lrr@{}}
\toprule
Source & Training samples & Test samples \\
\midrule
\multicolumn{3}{l}{Adapted visual sources} \\
EgoWalk\cite{ref22} & 296,899 & 2,997 \\
ADE20K\cite{ref29} & 159,053 & 2,998 \\
FPV Crosswalk\cite{ref30} & 33,973 & 2,247 \\
CARLA\cite{ref31} & 9,397 & 2,257 \\
RoomTour3D\cite{ref32} & 16,631 & 111 \\
\midrule
\multicolumn{3}{l}{Generated sources} \\
Simulated scenes & 110,224 & 1,632 \\
Photorealistic generation & 10,031 & 741 \\
Small Alley & 3,135 & 597 \\
\midrule
Total & 639,343 & 13,580 \\
\bottomrule
\end{tabular*}\par
\vspace{1mm}
\parbox{\columnwidth}{\footnotesize Samples denote agent--instruction planning tasks, not unique images. Training counts exclude the test instances listed in the final column.}
\end{table}
\section{Experiments}

The experiments first evaluate region, goal, and route feasibility predictions, followed by conditional path generation and full-pipeline evaluation. Ablations then examine the Trav. Rep. interface, pixel-level segmentation supervision, and joint optimization. Finally, synchronous and asynchronous execution are compared, and LaTraNav is deployed on wheeled and quadrupedal robots.

\subsection{Experimental Setup}

The training and test data are constructed as described in Section~\ref{sec:data}, with their source composition summarized in Table~\ref{tab:eval_data}.

Independent cascade denotes a training baseline in which the VLM and flow-matching planner are trained separately and connected at inference. LaTraNav and this baseline use the same training set, and their VLMs are initialized from UFO\cite{ref13} weights based on InternVL2.5-8B\cite{ref36}. The flow-matching planner uses a DINOv2 ViT-S/14\cite{ref24} image encoder and is initialized from pretrained flow-matching weights. Joint optimization uses AdamW with a learning rate of $2\times10^{-4}$, an effective batch size of 32, and two training epochs. The semantic and path loss weights are 0.25 and 1.0, respectively. For the independent cascade, the VLM is trained for two epochs and the flow-matching planner for 300 epochs.

All methods are evaluated on the common test set for the subtasks supported by their outputs. UFO\cite{ref13} and LISA\cite{ref12} are evaluated with their original weights for traversable-region prediction, and RoboPoint\cite{ref34} is used for goal localization. InternVL2.5-8B\cite{ref36} provides a zero-shot baseline for agent-conditioned route feasibility. Path comparisons include the vision-language path module of VAMOS\cite{ref9}, paths derived from VENTURA\cite{ref33} outputs, and Strict A* on predicted masks. VAMOS-VLM is evaluated with RGB and the navigation instruction, without its usual image-space goal input, to assess transfer to the instruction-conditioned task.

For the full-pipeline Strict A* comparison, the same region and goal predictions and path anchor as LaTraNav are used. A* searches the decoded mask toward the predicted goal, whereas LaTraNav's flow-matching planner uses Trav. Rep. Conditional planning is evaluated separately with ground-truth goals. Strict A* uses eight-connected search without diagonal crossings through obstacle corners or endpoint snapping. No path is returned when the path anchor or goal lies outside the mask, or when they are disconnected.

\begin{table}[t]
\caption{Agent- and language-conditioned traversable regions and goal localization}
\label{tab:semantic_tasks}
\centering
\scriptsize
\setlength{\tabcolsep}{3pt}
\textbf{(a) Agent- and language-conditioned traversable region}\par
\vspace{1mm}
\begin{tabular*}{\columnwidth}{@{\extracolsep{\fill}}lcc@{}}
\toprule
Method & IoU $\uparrow$ & Dice $\uparrow$ \\
\midrule
UFO\cite{ref13} & 0.311 & 0.372 \\
LISA\cite{ref12} & 0.332 & 0.395 \\
Independent cascade & 0.8864 & 0.9302 \\
LaTraNav (ours) & 0.8817 & 0.9275 \\
\bottomrule
\end{tabular*}\par
\vspace{2mm}
\textbf{(b) Goal localization}\par
\vspace{1mm}
\begin{tabular*}{\columnwidth}{@{\extracolsep{\fill}}lccc@{}}
\toprule
Method & Goal L2 $\downarrow$ & SR@0.05 $\uparrow$ & SR@0.15 $\uparrow$ \\
\midrule
RoboPoint\cite{ref34} & 0.312 & 5.4\% & 25.8\% \\
Independent cascade & 0.142 & 45.6\% & 79.1\% \\
LaTraNav (ours) & 0.155 & 42.3\% & 76.8\% \\
\bottomrule
\end{tabular*}\par
\vspace{1mm}
\parbox{\columnwidth}{\footnotesize (a) The same 13,093 samples with valid ground-truth masks are used. UFO and LISA use original weights, whereas the task models are trained on the proposed data. (b) All 11,768 ground-truth feasible samples are included. Missing goals receive an L2 error of $\sqrt{2}$ and count as SR failures.}
\end{table}

\begin{table}[t]
\caption{Agent-conditioned route feasibility}
\label{tab:agent_judge}
\centering
\scriptsize
\setlength{\tabcolsep}{3pt}
\textbf{(a) Overall classification}\par
\vspace{1mm}
\begin{tabular*}{\columnwidth}{@{\extracolsep{\fill}}lccc@{}}
\toprule
Method & Acc. $\uparrow$ & Macro-F1 $\uparrow$ & False-feasible $\downarrow$ \\
\midrule
InternVL2.5-8B\cite{ref36} & 88.37\% & 68.71\% & 65.89\% \\
Independent cascade & 96.73\% & 93.58\% & 0.00\% \\
LaTraNav (ours) & 96.40\% & 92.99\% & 0.00\% \\
\bottomrule
\end{tabular*}\par
\vspace{2mm}
\textbf{(b) Accuracy by agent}\par
\vspace{1mm}
\begin{tabular*}{\columnwidth}{@{\extracolsep{\fill}}lccc@{}}
\toprule
Method & \makecell{Humanoid\\Robot} & \makecell{Large Wheeled\\Robot} & \makecell{Small Wheeled\\Robot} \\
\midrule
InternVL2.5-8B\cite{ref36} & 97.25\% & 84.28\% & 83.34\% \\
Independent cascade & 99.98\% & 94.89\% & 95.23\% \\
LaTraNav (ours) & 99.85\% & 94.94\% & 94.31\% \\
\bottomrule
\end{tabular*}\par
\vspace{1mm}
\parbox{\columnwidth}{\footnotesize All methods use the same 13,580 samples. InternVL2.5-8B is evaluated without task-specific training. False-feasible rate is the proportion of the 1,812 ground-truth infeasible samples predicted feasible.}
\end{table}

\subsection{Evaluation Metrics}

Traversable-region prediction is measured by intersection over union (IoU) and Dice score on samples with ground-truth masks. Goal localization is evaluated on all 11,768 ground-truth feasible tasks. Goal L2 is the mean Euclidean error in normalized image coordinates $(u/W,v/H)$, with refusals and missing goal predictions assigned the unit-square maximum distance of $\sqrt{2}$. Goal success rate at threshold $\alpha$ (SR@$\alpha$) is the fraction of these tasks with a predicted goal error below $\alpha$, counting refusals and missing predictions as failures. Agent-conditioned route feasibility is evaluated on all 13,580 samples using accuracy, macro-F1, and false-feasible rate. The latter is the fraction of annotated infeasible routes predicted feasible.

Conditional planning uses ground-truth goals and reports average displacement error (ADE) on each method's $n_P$ generated paths. ADE is the mean distance between corresponding points on the resampled predicted and reference paths in normalized image coordinates. It is an auxiliary measure because multiple acceptable paths may reach the same goal. Full-pipeline evaluation uses predicted goals or directly generated paths, without supplying the ground-truth goal.

For each path, the path-violation rates $\mathrm{PVR}_{\rm pred}$ and $\mathrm{PVR}_{\rm ref}$ denote the proportions of predicted and reference points outside the agent-specific ground-truth mask. Excess violation is defined as $\Delta\mathrm{PVR}=\mathrm{PVR}_{\rm pred}-\mathrm{PVR}_{\rm ref}$ to account for reference-path violations associated with projection and occlusion. Its mean and the relative trajectory-violation rate at 0.05 (RelTVR@0.05), defined as the fraction with $\Delta\mathrm{PVR}>0.05$, are computed on each method's $n_V$ generated paths with ground-truth masks. These subsets may differ between methods.

No-path rate and Joint$_{\rm all}$ use all 11,768 ground-truth feasible tasks. No-path rate measures missing path outputs, including refusals, parse failures, and planning failures. Joint$_{\rm all}$ counts a task as successful when a path is produced, its endpoint error relative to the ground-truth goal is below 0.15, and $\Delta\mathrm{PVR}\leq0.05$. Missing paths count as failures. Thus, Joint$_{\rm all}$ measures full-pipeline success across the full feasible set, whereas violation metrics characterize generated paths. Unlike Goal SR, its endpoint criterion evaluates the path endpoint rather than the VLM-predicted goal.

\begin{table*}[t]
\caption{Full-set task completion and quality of generated paths}
\label{tab:planning_tasks}
\centering
\scriptsize
\setlength{\tabcolsep}{3pt}
\textbf{(a) Conditional planning with ground-truth goals}\par
\vspace{1mm}
\begin{tabular*}{\textwidth}{@{\extracolsep{\fill}}llccc@{}}
\toprule
& & All feasible tasks & \multicolumn{2}{c}{Generated paths only} \\
\cmidrule(lr){3-3}\cmidrule(lr){4-5}
Method & Traversability input & No-path rate $\downarrow$ & $n_P$ & ADE $\downarrow$ \\
\midrule
UFO mask + Strict A* & UFO mask & 74.05\% & 3,054 & 0.0230 \\
LaTraNav mask + Strict A* & LaTraNav mask & 22.94\% & 9,069 & 0.0256 \\
Independent cascade & Trav. Rep. & 3.77\% & 11,324 & 0.0313 \\
LaTraNav (ours) & Trav. Rep. & 4.16\% & 11,279 & 0.0151 \\
\bottomrule
\end{tabular*}\par
\vspace{2mm}
\textbf{(b) Full-pipeline evaluation with predicted goals or directly generated paths}\par
\vspace{1mm}
\begin{tabular*}{\textwidth}{@{\extracolsep{\fill}}lccccc@{}}
\toprule
& \multicolumn{2}{c}{All feasible tasks} & \multicolumn{3}{c}{Generated paths with GT masks} \\
\cmidrule(lr){2-3}\cmidrule(lr){4-6}
Method & Joint$_{\rm all}$ $\uparrow$ & No-path rate $\downarrow$ & $n_V$ & $\Delta$PVR $\downarrow$ & RelTVR@0.05 $\downarrow$ \\
\midrule
VAMOS-VLM\cite{ref9} & 3.1\% & 59.5\% & 4,735 & 0.362 & 66.0\% \\
VENTURA\cite{ref33} & 28.1\% & 2.0\% & 11,448 & 0.134 & 27.3\% \\
LaTraNav mask/goal + Strict A* & 51.6\% & 30.5\% & 8,118 & 0.025 & 11.9\% \\
Independent cascade & 57.70\% & 3.77\% & 11,242 & 0.0895 & 29.83\% \\
LaTraNav (ours) & 63.95\% & 4.16\% & 11,196 & 0.0544 & 20.92\% \\
\bottomrule
\end{tabular*}\par
\vspace{1mm}
\parbox{\textwidth}{\footnotesize All feasible-task rates use 11,768 samples. GT denotes ground truth. (a) Every method receives the GT goal. ADE uses each method's $n_P$ generated paths. (b) LaTraNav, the independent cascade, and Strict A* use predicted goals. VAMOS-VLM and VENTURA provide path outputs without a separate goal-localization stage. Violation metrics use each method's $n_V$ generated paths with GT masks. LaTraNav and the independent cascade omit a path only when the VLM refuses, so their no-path rates equal the refusal rates on feasible tasks.}
\end{table*}

\subsection{Region, Goal, and Route Feasibility Predictions}

Table~\ref{tab:semantic_tasks}(a) reports agent- and language-conditioned traversable-region prediction on the same 13,093 annotated samples. LaTraNav achieves an IoU of 0.8817, compared with 0.311 for UFO and 0.332 for LISA using their original weights. This comparison supports adapting language-guided segmentation to the proposed traversability task. The independent cascade attains a similar IoU of 0.8864, allowing the subsequent planning results to be considered alongside region accuracy.

On the full feasible set, LaTraNav achieves a Goal L2 of 0.155 and SR@0.15 of 76.8\%, compared with 0.312 and 25.8\% for RoboPoint in Table~\ref{tab:semantic_tasks}(b). The independent cascade obtains a lower Goal L2 of 0.142 and higher SR@0.15 of 79.1\%. These goal and region scores are considered together with full-pipeline success to assess whether the predictions support the resulting paths.

For agent-conditioned route feasibility in Table~\ref{tab:agent_judge}, LaTraNav achieves 96.40\% accuracy and 92.99\% macro-F1, compared with 88.37\% and 68.71\% for zero-shot InternVL2.5-8B. The false-feasible rate decreases from 65.89\% to 0.00\% on the 1,812 annotated infeasible tasks. Relative to this baseline, accuracy increases by 10.66 and 10.97 percentage points for large and small wheeled robots, respectively. These gains support task-specific training for recognizing agent-dependent route restrictions.

\begin{table}[t]
\caption{Structural and training ablations}
\label{tab:ablation}
\centering
\scriptsize
\setlength{\tabcolsep}{3pt}
\begin{tabular*}{\columnwidth}{@{\extracolsep{\fill}}p{0.43\columnwidth}ccc@{}}
\toprule
Setting & IoU $\uparrow$ & $\Delta$PVR$^\ddagger$ $\downarrow$ & Joint$_{\rm all}$ $\uparrow$ \\
\midrule
Without Trav. Rep. & 0.8802 & 0.1318 & 54.74\% \\
Explicit-mask planner & 0.8832 & 0.0731 & 59.83\% \\
Without segmentation supervision & 0.2631 & 0.1043 & 59.47\% \\
Independent cascade & 0.8864 & 0.0895 & 57.70\% \\
LaTraNav (ours) & 0.8817 & 0.0544 & 63.95\% \\
\bottomrule
\end{tabular*}\par
\vspace{1mm}
\parbox{\columnwidth}{\footnotesize $^\ddagger$ Mean $\Delta$PVR uses each setting's generated paths with GT masks. Joint$_{\rm all}$ uses all 11,768 ground-truth feasible tasks for every setting.}
\end{table}

\subsection{Conditional Planning and Full-Pipeline Evaluation}

With ground-truth goals in Table~\ref{tab:planning_tasks}(a), Strict A* produces paths for 3,054 tasks using UFO masks and 9,069 using LaTraNav masks. The independent cascade and LaTraNav produce paths for 11,324 and 11,279 tasks, respectively, with ADEs of 0.0313 and 0.0151 on their generated subsets. Supplying the ground-truth goal isolates conditional path generation from goal-prediction error. The output counts additionally show how explicit search depends on mask connectivity and endpoint inclusion.

Under the full-pipeline evaluation in Table~\ref{tab:planning_tasks}(b), VAMOS-VLM and VENTURA attain Joint$_{\rm all}$ scores of 3.1\% and 28.1\%, while LaTraNav reaches 63.95\%. Relative to the independent cascade at 57.70\%, LaTraNav gains 6.25 percentage points despite slightly lower region and goal scores. The planning improvement therefore does not follow from higher standalone region or goal accuracy, supporting the contribution of joint optimization to full-pipeline success.

Using LaTraNav's region and goal predictions, Strict A* searches the decoded mask and reaches 51.6\% Joint$_{\rm all}$, with no path for 30.5\% of feasible tasks. The flow-matching planner conditioned on Trav. Rep. reaches 63.95\%, a gain of 12.35 percentage points. Strict A* records lower excess violation on the paths it generates, while its lower full-pipeline success reflects failures to produce paths under disconnected masks or out-of-mask goals or anchors. Generated-path quality and full-pipeline success thus describe complementary aspects of planning performance.

\begin{figure}[!t]
    \centering
    \includegraphics[width=\columnwidth]{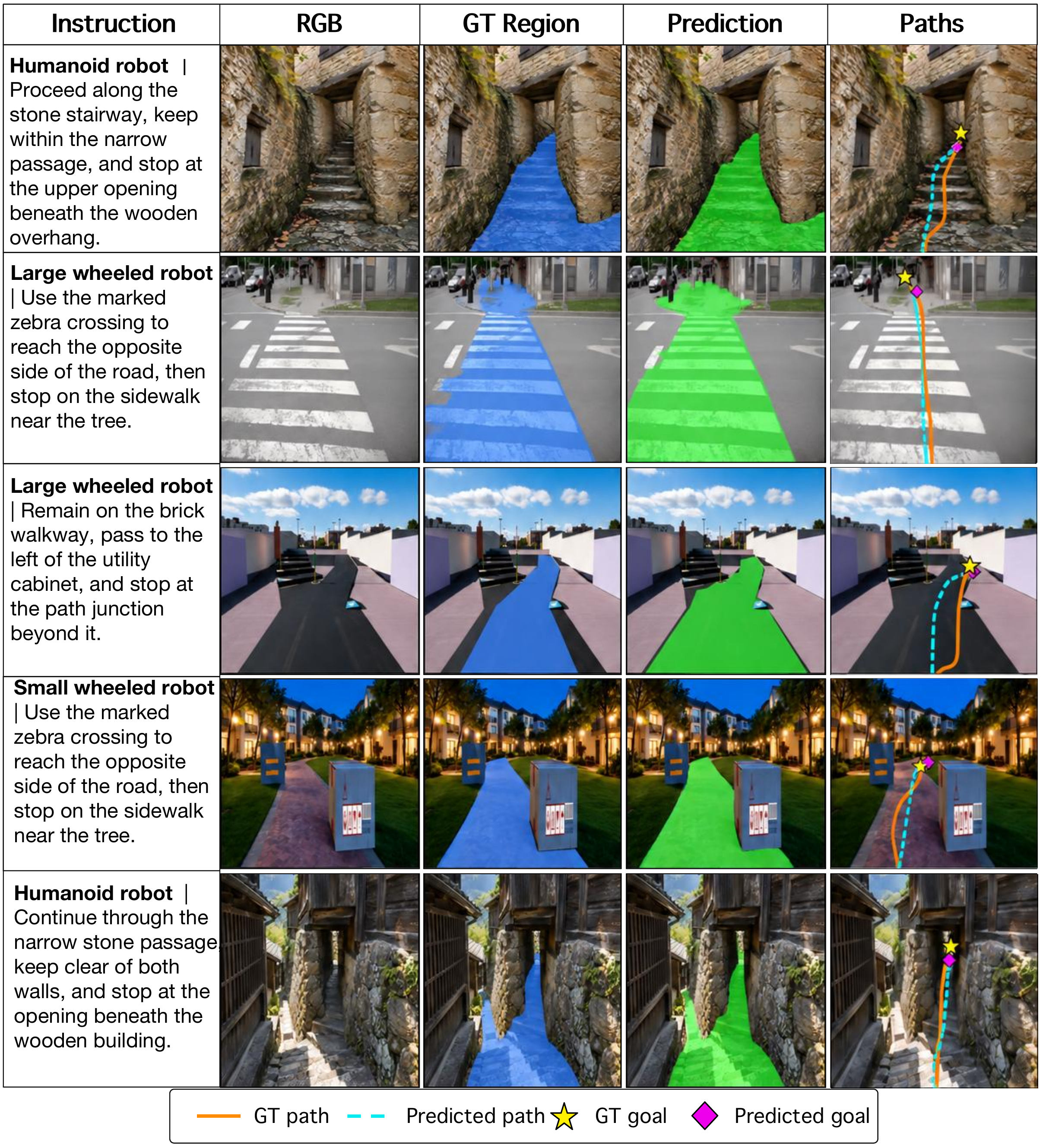}
    \caption{Qualitative results. Columns show the agent and instruction, RGB image, ground-truth region, predicted region, and paths. The shared legend identifies reference and predicted paths and goals.}
    \label{fig:qualitative_results}
\end{figure}

Fig.~\ref{fig:qualitative_results} visualizes region and goal predictions together with generated paths across crossings, ramps, nighttime scenes, and narrow passages.

\subsection{Structural and Training Ablations}

Table~\ref{tab:ablation} examines the Trav. Rep. interface, pixel-level segmentation supervision, and joint optimization.

When Trav. Rep. is withheld from the flow-matching planner, the VLM and mask decoder are retained. Mask IoU remains at 0.8802, but Joint$_{\rm all}$ drops to 54.74\%, showing that accurate segmentation alone is insufficient when agent- and language-conditioned traversability is unavailable to path generation. The explicit-mask planner conditions the flow-matching planner on the decoded mask and reaches 59.83\%. LaTraNav achieves 63.95\% at a similar mask IoU, improving success by 4.12 percentage points and reducing mean $\Delta$PVR from 0.0731 to 0.0544. This comparison supports the learned latent interface for translating region predictions into paths.

Removing pixel-level segmentation supervision reduces IoU to 0.2631 and Joint$_{\rm all}$ to 59.47\%, supporting its role in grounding the Trav. Rep.

Compared with the independent cascade, LaTraNav reduces mean $\Delta$PVR from 0.0895 to 0.0544 alongside the 6.25-point gain in Joint$_{\rm all}$. These ablations support the Trav. Rep. interface, pixel-level segmentation supervision, and joint optimization as complementary components of the planning system.

\subsection{Asynchronous Execution Design}

Synchronous execution pairs each VLM inference with one path-generation call. Asynchronous execution reuses the cached goal and Trav. Rep. for path generation while the next VLM inference runs concurrently.

\begin{table}[!t]
\caption{Measured synchronous and asynchronous execution}
\label{tab:async_execution}
\centering
\scriptsize
\setlength{\tabcolsep}{3pt}
\begin{tabular*}{\columnwidth}{@{\extracolsep{\fill}}lcccc@{}}
\toprule
Execution & \makecell{Path updates\\(Hz) $\uparrow$} & \makecell{Planner latency\\(ms) $\downarrow$} & \makecell{Semantic updates\\(Hz)} & \makecell{Real-robot\\success rate $\uparrow$} \\
\midrule
Synchronous & 1.21 & 63.0 & 1.21 & 0.96 \\
Asynchronous & 7.32 & 86.0 & 1.21 & 0.94 \\
\bottomrule
\end{tabular*}\par
\vspace{1mm}
\parbox{\columnwidth}{\footnotesize Rates count completed outputs per second. Planner latency is the mean flow-matching planner inference time, excluding VLM inference.}
\end{table}

Table~\ref{tab:async_execution} shows a $6.05\times$ increase in path-update rate, from 1.21 to 7.32 Hz, with the semantic-update rate unchanged at 1.21 Hz. The resulting semantic-update interval is approximately 0.83 s, during which the flow-matching planner combines the cached goal and Trav. Rep. with current visual features. Planner latency rises from 63 to 86 ms, but semantic reuse still enables multiple path updates between semantic updates.

\subsection{Real-Robot Deployment}

LaTraNav is deployed on the wheeled AgileX RANGER MINI 3.0 and quadrupedal UNIUBI Cyvet, whose traversability condition uses the humanoid-robot capability profile. Forward-facing RGB images from each robot platform are streamed to an external workstation equipped with an NVIDIA A100-SXM4-80GB GPU, where the VLM and flow-matching planner run. Generated image-space paths are converted into robot-frame waypoints using camera calibration and tracked by the platform's model predictive control (MPC) controller.

\section{Conclusion}

We presented LaTraNav, a dual-system framework that couples a slow VLM with a flow-matching planner through agent- and language-conditioned traversability representations. Pixel-level segmentation supervision and path supervision jointly ground Trav. Rep. in spatial semantics and adapt it to robot capabilities and task requirements. A controllable data-construction pipeline combining a simulation branch, photorealistic image generation, and multi-source visual data provides aligned training supervision. Experiments demonstrate improved full-pipeline success over explicit-mask conditioning and more frequent path updates under asynchronous execution, supporting Trav. Rep. as a reusable interface between language understanding and adaptive visual navigation.

\bibliographystyle{IEEEtran}
\bibliography{references}

\end{document}